# CID: Measuring Cognitive Work Through Semantic State Change

Ron Fulbright

University of South Carolina Upstate
Spartanburg, SC USA
fulbrigh@uscupstate.edu

**Abstract**
Traditional information theory provides powerful tools for analyzing statistical properties of information but offers limited insight into meaning and semantic transformation. It has never been possible to quantify the meaning of an arbitrary piece of information. Large language models represent information as vectors embedded within high-dimensional semantic spaces, enabling meaning to be analyzed geometrically. This paper defines semantic state as the vector representation of information within a model's semantic space and defines cognitive work as the transformation between semantic states—movement through semantic space. To quantify this transformation, we introduce the Contextual Information Differential (CID), a covariance-aware metric based on Mahalanobis distance measuring semantic displacement relative to the structure of the embedding manifold. By treating meaning as a geometric property of semantic space, CID provides a practical method for measuring semantic change, cognitive transformation, and information processing. This framework establishes a foundation for analyzing cognitive processing, learning, reasoning, knowledge generation, synthetic expertise, and human–AI collaboration through implementation-independent measurement of semantic state change in both human and artificial systems.

## 1 INTRODUCTION

While traditional information theory (Shannon, 1948) excels at quantifying statistical properties of information, it provides limited means for measuring meaning and changes in meaning itself—semantics. Recent advances in artificial intelligence (AI) suggest a solution. Modern Large Language Models (LLMs) learn distributed representations in which concepts, sentences, and documents are represented as vectors embedded within a high-dimensional semantic space (Bengio et al., 2003; Mikolov et al., 2013; Vaswani et al., 2017). These representations capture semantic relationships geometrically, allowing similar concepts to occupy nearby regions of semantic space while dissimilar concepts are separated by greater distances.

The transition from symbolic representations to distributed vector representations represents a fundamental shift in how information and cognitive processing can be analyzed. In embedding-based semantic spaces, information is no longer represented solely by words or symbols but by geometric relationships within a learned semantic manifold. Any collection of tokens (words, sub-words, sentences, paragraphs, documents, etc.) corresponds to a single embedding vector in the semantic space. Each vector occupies a point in the multi-dimensional semantic space representing the information's semantic state. When information is changed the new vector represents a different semantic state. The resulting displacement reflects a change in meaning rather than merely a change in textual form. Semantic transformation can be viewed as movement through semantic space.

This observation leads naturally to the concept of cognitive work. We define cognitive work, $W$, as the semantic transformation required to move information from an input semantic state to an output semantic state:

$$W = E_{out} - E_{in} \tag{1}$$

where $E_{in}$ represents the input semantic state and $E_{out}$ represents the semantic state following transformation, and $W$ is the semantic transformation vector. Under this formulation, all cognitive processes perform cognitive work because each operation changes the semantic state of the information involved.

Cognitive work is displacement within semantic space. However, raw displacement alone is insufficient for measuring meaningful semantic change. Modern embedding spaces are highly anisotropic, meaning different directions exhibit different variances and correlations (Ethayarajh, 2019). Movement through a dense region of semantic space may represent a routine transformation, while an equivalent geometric movement through a sparse region may represent substantial conceptual novelty. To account for these differences, this paper introduces the Contextual Information Differential (CID), a covariance-aware measure of semantic transformation based on Mahalanobis distance (Mahalanobis, 1936). Unlike other distance measures, Mahalanobis distance incorporates the covariance structure of the underlying space allowing displacement to be measured relative to the geometry of the semantic manifold rather than raw coordinate distance. CID is formally defined as:

$$CID = \sqrt{(E_{out} - E_{in})^T \, \Sigma^{-1} \, (E_{out} - E_{in})} \tag{2}$$

where Σ represents the covariance structure of the embedding manifold, $T$ means transpose, $E_{in}$ represents the input semantic state of the corresponding information $S_{in}$ prior to transformation and $E_{out}$ represents the semantic state of the corresponding information, $S_{out}$ following transformation. By incorporating covariance information, CID measures not merely how far information moves, but how significant the movement is relative to the learned structure of semantic space. In this sense, CID can be interpreted as a measure of the cognitive work performed during a transformation of information. The framework developed in this paper rests on three foundational claims:

1. Information possesses a measurable semantic state.
2. Cognitive work corresponds to transformations between semantic states.
3. CID provides a quantitative measure of the magnitude of the transformation.

## 2 RELATED WORK

The Contextual Information Differential (CID) framework draws upon research from information theory, distributed semantic representations, geometric approaches to meaning, and multivariate distance measurement. While these fields have provided powerful methods for representing and comparing information, none directly address the problem of measuring semantic transformation as a form of cognitive work. CID seeks to bridge this gap by combining ideas from these disciplines into a unified framework for quantifying semantic state change.

### 2.1 Information Theory

Modern information science is rooted in Shannon's foundational work on communication theory (Shannon, 1948). Shannon introduced mathematical measures for quantifying uncertainty, entropy, and information transmission. However, Shannon explicitly noted the semantic aspects of communication were outside the scope of the theory. As a result, traditional information theory provides powerful tools for measuring information quantity but offers limited mechanisms for measuring changes in meaning. The CID framework builds upon information theory while addressing a fundamentally different question. Rather than asking how much information is transmitted, CID asks how much the semantic state of information changes during transformation.

### 2.2 Distributed Semantics

The emergence of neural language models enabled information to be represented in continuous vector spaces. Bengio et al. (2003) demonstrated neural networks could learn distributed representations capturing semantic relationships among words. Subsequent work by Mikolov et al. (2013) showed word embeddings encode meaningful semantic and syntactic relationships through geometric structure.

More recently, transformer-based architectures have produced increasingly sophisticated embedding spaces capable of representing complex semantic relationships among concepts, documents, and knowledge structures (Vaswani et al., 2017; Devlin et al., 2019). These developments provide the foundation for treating meaning as a geometric property of a learned semantic manifold.

CID builds directly upon these advances by treating embedding vectors as representations of semantic state. However, unlike prior approaches that primarily use embeddings to measure similarity, CID focuses on measuring semantic transformation.

### 2.3 Semantic Similarity Metrics

Several distance and similarity measures are commonly used to compare embedding vectors. Cosine similarity measures the angular separation between vectors and has become one of the most widely used measures of semantic similarity in vector space models (Salton et al., 1975). Euclidean distance measures straight-line geometric separation, while Manhattan distance measures the sum of coordinate-wise differences. Dot-product similarity evaluates vector alignment and is commonly used in retrieval and recommendation systems.

While these measures are effective for determining whether two representations are similar, they do not explicitly measure the contextual significance of semantic change. Two semantic transformations may produce identical cosine similarities or Euclidean distances while differing substantially in conceptual importance. CID differs from these approaches in that it is designed to quantify semantic transformation rather than semantic similarity.

### 2.4 Mahalanobis Distance

The mathematical foundation of CID derives from the Mahalanobis distance introduced by Mahalanobis (1936). Unlike Euclidean distance, Mahalanobis distance incorporates the covariance structure of the underlying space, allowing displacement to be measured relative to local variance and dimensional correlations. This property makes Mahalanobis distance particularly well suited for embedding spaces, which are known to exhibit anisotropic structure, dimensional dependencies, and non-uniform semantic density (Ethayarajh, 2019). CID extends this principle by applying covariance-aware distance measurement to semantic state transformations within AI semantic spaces.

### 2.5 Cognitive Load and Cognitive Work

The concept of cognitive work has been studied extensively within cognitive psychology, human factors, and educational research. Cognitive Load Theory (Sweller, 1988) focuses on the mental effort required to perform learning tasks and distinguishes between intrinsic, extraneous, and germane cognitive load. Similarly, research in human factors has examined mental workload as a measure of task difficulty and resource utilization.

CID differs from these approaches by focusing not on the effort required to perform a task but on the magnitude of the resulting semantic transformation. In this sense, traditional cognitive load measures the resources expended during cognition, whereas CID measures the semantic change produced by cognition.

### 2.6 Distinction From Prior Work

The central distinction between CID and prior approaches can be summarized as follows:

- Traditional information theory measures uncertainty and information transmission.
- Embedding similarity metrics measure semantic proximity.
- Cognitive load theories measure mental effort.
- CID measures semantic transformation.

By representing information as semantic states and cognitive activity as transformations between those states, CID provides a framework for quantifying cognitive work as contextual semantic displacement. To our knowledge, no existing approach explicitly combines semantic state representation, geometric transformation, and covariance-aware measurement into a unified framework for measuring semantic change. This distinction constitutes the primary contribution of the present work.

## 3 SEMANTIC STATE

### 3.1 Information as Semantic State

LLMs encode information as points within a high-dimensional semantic space learned from vast collections of text (Bengio et al., 2003; Mikolov et al., 2013). During training, LLMs learn statistical regularities from massive corpora of human knowledge and language. These regularities become embedded within a geometric structure capturing semantic relationships among entities, concepts, events, and abstractions (Vaswani et al., 2017). Information is broken into tokens and transformed into embedding vectors, $E$, where each vector represents a point within a high-dimensional semantic space:

$$E \in \mathbb{R}^d \tag{3}$$

where $d$ represents the dimensionality of the embedding space. In modern embedding systems, $d$ often ranges from hundreds to thousands of dimensions. Each embedding vector consists of $d$ floating point numbers and one can think of these values as being the address of the vector's point in semantic space. Although these dimensions generally lack direct human interpretation, their collective structure captures complex semantic relationships among concepts, facts, and ideas. The resulting semantic space possesses several important properties:

- Similar concepts occupy nearby regions.
- Dissimilar concepts occupy distant regions.
- Conceptual relationships become geometric relationships.
- Semantic transformations correspond to movement through the space.

For example, concepts associated with medicine tend to cluster together, while concepts associated with finance occupy different semantic neighborhoods. Likewise, variations of the same idea occupy nearby locations within the space. The position occupied by a vector within this space constitutes its semantic state, allowing meaning to be represented as a measurable geometric property rather than merely a sequence of symbols.

Since similar topics tend to occupy nearby regions in semantic space, semantic similarity can be interpreted geometrically. More importantly, semantic neighborhoods establish local context. The meaning of a semantic state depends not only on its position but also on its relationship to surrounding states. This local structure forms the basis for understanding semantics and semantic transformation.

### 3.2 Semantic Transformations

Information rarely remains static. Humans, AI systems, computer software, even natural processes routinely transform information. Here, we call transformation of information a *cognitive process* as depicted in Figure 1 (Fulbright, 2020).

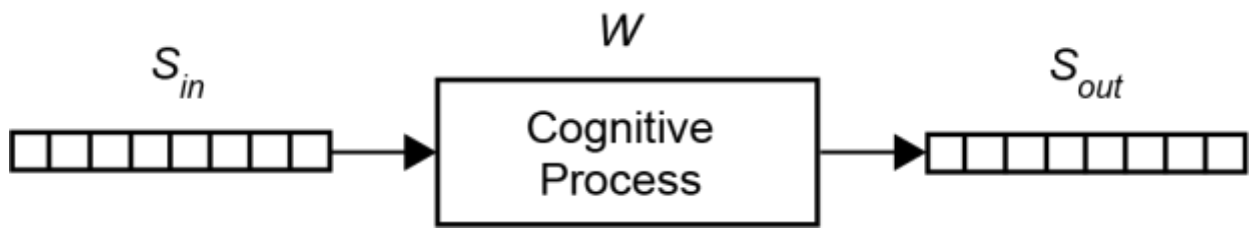

Figure 1: A cognitive process transforms information from an input state to an output state by performing an amount of cognitive work.

where $S_{in}$ and $S_{out}$ is the information before and after the transformation and $W$ is the cognitive work represented by the transformation. Since $S_{in}$ and $S_{out}$ can be represented as corresponding embedding vectors $E_{in}$ and $E_{out}$ respectively, the distance between $E_{in}$ and $E_{out}$ is the contextual information differential. CID, calculated using Eq. 2, and is a measurement of movement through semantic space quantifying the cognitive work associated with the transformation effected by a cognitive process.

The semantic significance of these transformations depends not only on how far information moves but also on the structure of the semantic manifold through which it moves. Thus, information processing in general can be viewed as a sequence of semantic state transitions:

$$S1 \rightarrow S2 \rightarrow S3 \rightarrow \cdots \quad \text{corresponding to} \quad E1 \rightarrow E2 \rightarrow E3 \rightarrow \cdots$$

Furthermore, in this framework, it does not matter who or what is performing the transformation. Cognitive work can be performed by either a human or an artificial system, or both. CID is an implementation-independent metric for cognitive work.

## 4 COGNITIVE WORK

### 4.1 Cognitive Work as Semantic Displacement

By analogy, the concept of work has a long history in physics, where work represents the displacement of an object through a space under the influence of force. Information occupies a location within semantic space. Cognitive processes act upon information and move it to a new location. The resulting displacement represents the work performed by the cognitive system. Some cognitive processes result in very little semantic displacement while others result in very large displacements. For example:

- A paraphrase may produce relatively little semantic displacement because the meaning remains largely unchanged.
- A summary may produce greater displacement because information is compressed and reorganized.
- An explanation may produce even larger displacement by introducing conceptual structure and contextual information.
- A scientific discovery may produce substantial displacement by generating previously unknown semantic relationships.

Consider the statement: $S_{in}$ = "The Earth revolves around the Sun." and a simple paraphrase transformation such as: $S_{out}$ = "The Earth orbits the Sun." This transformation produces only a modest semantic displacement because the underlying meaning changes very little (a small CID). However, consider the transformation to $S_{out}$ = "The Earth follows an elliptical orbit around the Sun due to gravitational attraction." This transformation introduces additional concepts, causal relationships, and contextual structure. The resulting semantic displacement is larger (a larger CID). A more advanced transformation might connect the statement to celestial mechanics, orbital dynamics, and astrophysical theory (an even larger CID). In each case, a cognitive process modifies meaning and therefore changes semantic state. CID measures the cognitive work.

### 4.2 Why Other Distance Metrics Are Insufficient

Calculating distances in semantic spaces is not new. AI researchers currently use: cosine distance, Euclidean distance, dot product distance, and Manhattan distance.

Modern embedding spaces are highly anisotropic (Ethayarajh, 2019). Some directions exhibit high variance and correspond to common semantic transformations. Other directions exhibit low variance and may represent rare or conceptually important changes. Equal geometric displacement can therefore correspond to dramatically different levels of semantic significance. For example, movement within a dense semantic neighborhood may represent a common reformulation, while an equivalent displacement into a sparse region may represent a novel insight. Consequently, cognitive work must be measured relative to the structure of the semantic manifold itself.

CID offers several advantages over cosine similarity as a measure of semantic transformation. Cosine similarity measures only the angular separation between two vectors and is primarily designed to assess semantic similarity and ignores the statistical structure of the embedding space, treating all directions as equally significant (Salton & Wang, 1975).

$$Cosine(A, B) = 1 - \frac{(A \cdot B)}{(||A||\ ||B||)} \tag{4}$$

In contrast, CID incorporates the covariance structure of the semantic space through the inverse covariance matrix, allowing semantic displacement to be evaluated relative to the geometry of the space. As a result, CID measures not merely whether two semantic states are different, but how significant the difference is. Whereas cosine similarity answers the question "How similar are these two semantic states?" CID answers the question "How much meaningful transformation occurred between them?"

Euclidean distance also treats all dimensions of the embedding space as independent and equally important.

$$Euclidean(A, B) = \sqrt{\sum_{i=1}^{n}(A_i - B_i)^2} \tag{5}$$

As a result, identical Euclidean displacements may represent dramatically different levels of semantic significance. CID addresses this limitation by incorporating the covariance structure of the semantic manifold through Mahalanobis distance, weighting semantic displacement according to the geometry of the underlying semantic space.

Dot products are strongly influenced by vector magnitude and primarily measure alignment rather than semantic significance.

$$A \cdot B = \sum_{i=1}^{n} A_i B_i \tag{6}$$

Two vectors may produce similar dot products despite representing transformations of very different semantic importance. Furthermore, dot products ignore the statistical structure of the embedding space and treat all dimensions uniformly. Whereas dot-product similarity measures vector alignment, CID measures covariance-adjusted semantic transformation.

Manhattan distance measures semantic displacement as the sum of coordinate-by-coordinate differences and assumes all dimensions contribute independently and equally to the transformation.

$$Manhattan(A, B) = \sum_{i=1}^{n}|A_i - B_i| \tag{7}$$

As a result, Manhattan distance may overestimate or underestimate the importance of a semantic transformation by ignoring the underlying structure of the embedding space. CID addresses this limitation by incorporating the covariance structure of the semantic space through Mahalanobis distance, allowing semantic displacement to be evaluated relative to the learned geometry of the semantic space. Consequently, CID measures the contextual significance of semantic change rather than simply the aggregate magnitude of coordinate differences.

### 4.3 Examples of Cognitive Processes and Cognitive Work

*Translation*

Consider the English sentence "Water freezes at zero degrees Celsius." translated into Spanish: "El agua se congela a cero grados Celsius." The wording is completely different, yet the underlying meaning remains essentially unchanged. An embedding model capable of multilingual semantic representation should place these sentences in nearby semantic regions despite being written in different languages (Devlin et al., 2019). Consequently, CID is low. In fact, the goal of translation should be CID=0. This example highlights an important property of semantic-state analysis: CID measures changes in meaning rather than changes in surface form.

*Paraphrasing*

Consider the sentence: "The cat sat on the mat." A paraphrased version might be: "The cat rested on the mat." Although the wording changed, the underlying meaning remains largely unchanged. Both sentences describe the same entities, actions, and relationships. As a result, their semantic states occupy nearby positions within semantic space, therefore CID is small meaning paraphrasing requires a small amount of cognitive work.

*Summarization*

Consider the following text: "The company reported strong quarterly earnings, expanded operations into three new international markets, and announced plans to increase research and development spending over the next fiscal year." A summary might be: "The company reported strong growth and expansion plans." The summary preserves the central message while removing detail, specificity, and contextual information. Although the summary remains semantically related to the original text, information has been compressed and reorganized. Consequently, the output semantic state differs more substantially from the input state than in the paraphrasing example. Summarization therefore performs more cognitive work than paraphrasing because meaningful semantic restructuring has occurred therefore CID is larger.

*Explanation*

Consider the statement: "Photosynthesis converts sunlight into chemical energy." An explanatory transformation might produce: "Photosynthesis enables plants to capture energy from sunlight and store that energy in the form of glucose through a series of biochemical reactions." Unlike summarization, explanation introduces additional concepts, causal relationships, and contextual information. The output semantic state now contains information not explicitly present in the original statement reflecting the addition of new semantic structure rather than simple compression or reformulation. Therefore, CID is larger still.

*Conceptual Synthesis*

Consider two separate observations: "Exercise improves cardiovascular health" and "Exercise reduces symptoms of depression." A synthesized interpretation might be: "Regular exercise simultaneously benefits physical and mental health, making it a powerful intervention for overall human well-being." Unlike summarization or explanation, synthesis combines multiple ideas into a more generalized conceptual framework. The output semantic state represents a newly organized conceptual structure rather than a direct reformulation of the input information. Consequently, the greater CID value reflects the creation of new semantic relationships among previously separate concepts.

*Scientific Discovery*

Scientific discovery represents one of the most significant forms of semantic transformation. Consider historical examples such as: Newton's formulation of classical mechanics, Darwin's theory of evolution, Einstein's theory of relativity. In each case, existing observations were transformed into entirely new explanatory frameworks. These transformations did not merely reorganize information. They generated new semantic structures altering how large bodies of knowledge were interpreted thereby representing very large CID.

# 5 APPLICATIONS

## 5.1 Cognitive Work in Humans and Artificial Systems

Cognitive work is performed regardless of whether the transformation is performed by a human or an artificial intelligence system. A human teacher explaining a concept, a scientist generating a hypothesis, or an LLM producing a summary all perform semantic transformations. In each case, information moves from one semantic state to another by virtue of the cognitive processing performed. The semantic-state framework and CID, therefore, provides a common language for examining cognitive activity across both biological and artificial cognitive systems. This perspective is particularly relevant as AI systems increasingly engage in human-level cognitive processes. CID provides an implementation-independent mechanism for quantifying the semantic significance of information transformations within a unified framework.

## 5.2 Information Content and Semantic Content

Traditional information theory focuses primarily on uncertainty and statistical characteristics (Shannon, 1948). Shannon information is measured in terms of probabilities, entropy, and coding efficiency. CID addresses a different question: How much did the meaning change? Two documents may exhibit identical Shannon information while possessing dramatically different semantic content. Conversely, a profound conceptual insight may require only a small amount of additional textual information while producing a large semantic transformation. CID therefore complements rather than replaces traditional information measures. Shannon measures information content while CID measures semantic content. It has never been possible to quantify the value of an arbitrary piece of information. This has been because there has been no simple way to measure semantic impact. However, now, the semantic spaces associated with LLMs gives us a "semantic standard" to measure arbitrary information against and CID gives us the easy-to-calculate metric.

## 5.3 Education

CID gives us a way to assess a student's answer relative to a standard not currently possible. Imagine a question asking a student to explain a certain concept. The concept is represented as a point in semantic space. When a student answers the question, CID can calculate the semantic difference between the answer and the concept. The goal of teaching is then to instruct the student until their answers close the distance to the standard representation of the concept to within a certain tolerance. Not only can CID be used to monitor student progress over time, but CID can also be used to automatically grade student responses. Interestingly, the standard conceptual representation in semantic space can be used universally so every student would be graded against the same standard. This promises to enhance the quality of education in every grade and every school.

## 5.4 Cognitive Psychology

The field of cognitive psychology studies of cognitive load, mental workload, and the human cognitive architecture. Critical in this field is the ability to measure the mental effort required to perform a task—a measure of how demanding a task is on human cognition. Traditional cognitive load research asks: "How much mental effort is required to complete a task?" CID asks: "How much semantic transformation occurred as a result of the task?" Those are related but different concepts. Cognitive Load Theory measures the effort required to transform information, whereas CID measures the magnitude of the semantic transformation itself. This distinction positions CID with interesting possible applications within existing cognitive-load literature.

### 5.5 Artificial Intelligence

The emergence of large-scale embedding models creates an unprecedented opportunity to measure semantic transformation directly. Because modern AI systems already represent information geometrically, CID can potentially be applied to:

- Reasoning systems
- Question answering
- Summarization
- Educational systems
- Scientific discovery
- Synthetic expertise
- Agent evaluation
- Human-AI collaboration

In each case, semantic state transitions become measurable through covariance-aware semantic displacement. CID therefore provides a foundation for quantifying cognitive work within both biological and artificial cognitive systems. CID has the potential to play an important role in modern generative AI systems by providing a quantitative measure of semantic transformation during content generation. Current generative models are typically evaluated using metrics such as perplexity, task accuracy, or human preference ratings, which provide limited insight into how much meaning has changed during a generation process. By measuring the contextual semantic displacement between an input state and an output state, CID offers a mechanism for assessing the magnitude of reasoning, explanation, summarization, creativity, or knowledge generation performed by the model. This capability could support the evaluation of agentic AI systems, synthetic expertise, autonomous knowledge generation, and human-AI collaboration by quantifying not merely what was produced, but how significantly the underlying semantic state was transformed.

### 5.6 Cognitive Work in Attention Layers

In modern transformer architectures, each attention layer incorporates additional contextual information as the representation moves through the layers $L_0 \rightarrow L_1 \rightarrow L_2 \rightarrow \cdots \rightarrow L_n$ (Vaswani et al., 2017). Each attention layer transforms the representation thereby performing cognitive work as given by:

$$CID_i = \sqrt{(E_{i+1} - E_i)^T \, \Sigma^{-1} \, (E_{i+1} - E_i)} \quad (8)$$

where $E$ is the evolving representation at attention layer $i$. The total amount of cognitive work performed by one pass through $n$ attention layers is:

$$CID_{total} = \sum_{i=1}^{n} CID_i. \quad (9)$$

### 5.7 Internal Cognitive Work of Multi-Step Reasoning

Often, performing a cognitive task requires a lot of intermediate processing. Imagine a question answered by a human or a generative AI. The net cognitive work can be calculated by comparing the question with the answer:

$$CID_{net} = (E_{question} \rightarrow E_{answer}). \tag{10}$$

One gets a very different view of the cognitive work involved in the task when the cognitive work of each intermediate step is calculated. Imagine there were five major processing steps involved to answer the question. The cognitive work of the entire path is:

$$CID_{path} = CID_{step\ 1} + CID_{step\ 2} + CID_{step\ 3} + CID_{step\ 4} + CID_{step\ 5} \tag{11}$$

Which can be a very different answer than $\boldsymbol{CID_{net}}$. In fact, this suggests ideas like comparing the two:

$$\frac{CID_{path}}{CID_{net}} \tag{12}$$

measuring something we might call reasoning depth, cognitive effort, reasoning complexity, or semantic complexity. A trivial process would have $CID_{path} \approx CID_{net}$ while a difficult reasoning problem might have $CID_{path} \gg CID_{net}$.

## 6 LIMITATIONS

Several limitations of the proposed CID framework should be acknowledged. First, CID is dependent upon the underlying embedding model and semantic space. Different language models may generate different semantic geometries, covariance structures, and embedding representations for the same information. Consequently, CID values obtained from different models may not be directly comparable.

Second, CID measures semantic transformation, not correctness, factual accuracy, logical validity, or truth. A transformation may produce a high CID value while still containing factual errors or flawed reasoning. CID quantifies the magnitude of semantic change rather than the quality of the resulting output.

Third, the framework assumes that semantic states can be adequately represented within high-dimensional embedding spaces. While modern language models have demonstrated remarkable success using distributed semantic representations, the extent to which embedding vectors capture all aspects of meaning remains an active area of research.

Fourth, the current paper presents CID as a conceptual and mathematical framework. Although illustrative examples are provided, empirical validation remains future work. Additional research is needed to evaluate the behavior of CID across different models, domains, tasks, and transformation types.

Finally, the interpretation of CID as a measure of cognitive work should be viewed as a theoretical construct rather than a direct measurement of human cognition. CID measures semantic-state transformation within a representational space and may serve as an approximation of cognitive activity, but it does not claim to directly measure neurological processes or mental effort.

Despite these limitations, CID provides a foundation for investigating semantic transformation as a measurable phenomenon and suggests new opportunities for studying learning, reasoning, knowledge generation, and human-AI collaboration through the geometry of semantic state change.

## 7 CONCLUSION

Modern AI makes it possible to represent meaning geometrically within semantic space. This paper introduced semantic state as a representation of meaning, cognitive work as the transformation between semantic states, and the Contextual Information Differential (CID) as a covariance-aware measure of semantic transformation. Unlike traditional similarity metrics, CID measures the contextual significance of semantic change relative to the geometry of the underlying embedding manifold. As AI systems increasingly participate in higher-level cognitive work, measures such as CID establish a foundation for the scientific study of semantic change.

### AI Use Disclosure

Generative AI was used as an auxiliary tool in the development of this paper. Its use included assisting with conceptual exploration, evaluating alternative explanations, refining technical descriptions, and improving the clarity and organization of the manuscript. All AI-generated suggestions and content were critically reviewed, evaluated, and edited by the author. The author retains full responsibility for the originality, technical accuracy, research contributions, interpretations, and conclusions presented in this paper.